\documentclass[conference]{IEEEtran}
\IEEEoverridecommandlockouts
\usepackage{cite}
\usepackage{amsmath,amssymb,amsfonts}
\usepackage{algorithmic}
\usepackage{graphicx}
\usepackage{textcomp}
\usepackage{xcolor}
\usepackage{hyperref}
\usepackage[T1]{fontenc}
\newcommand{\englsk}[2]{(engl.~\emph{#1}, #2)}
\def\BibTeX{{\rm B\kern-.05em{\sc i\kern-.025em b}\kern-.08em
    T\kern-.1667em\lower.7ex\hbox{E}\kern-.125emX}}

\renewcommand\refname{Literatura}
\renewcommand{\abstractname}{Sažetak}
\renewcommand\tablename{TABLICA}
\begin{document}

\title{Semanti\v{c}ka Segmentacija Oblaka To\v{c}aka}

\author{\IEEEauthorblockN{Ivan Martinović}
\IEEEauthorblockA{\textit{Sveu\v{c}ilište u Zagrebu, Fakultet elektrotehnike i ra\v{c}unarstva} \\
ivan.martinovic@fer.hr}
}

\maketitle

\begin{abstract}
Semantička segmentacija važan je i dobro poznati zadatak u području računalnoga vida u kojem svakom elementu ulaza pokušavamo dodijeliti pripadajući semantički razred. Govorimo li o semantičkoj segmentaciji 2D slika, elementi ulaza su pikseli. S druge strane, ulaz može biti i oblak točaka, dok jedan element ulaza predstavlja jednu točku ulaznog oblaka točaka. Pod pojmom oblak točaka podrazumijevamo skup točaka definiranih s prostornim koordinatama s obzirom na neki referentni koordinatni sustav. Osim samog položaja točaka u prostoru, za svaku točku mogu biti definirane i druge značajke, kao npr.~RGB komponente. U okviru ovoga rada semantičku segmentaciju proveli smo nad skupom podataka S3DIS, pri čemu svaki oblak točaka predstavlja jednu prostoriju. Na skupu podataka S3DIS naučili smo modele: \emph{PointCNN}, \emph{PointNet++}, \emph{Cylinder3D}, \emph{Point Transformer} i \emph{RepSurf}. Dobivene rezultate usporedili smo s obzirom na standardne evaluacijske metrike za semantičku segmentaciju, a prikazali smo i usporedbu modela s obzirom na brzinu zaključivanja.
\end{abstract}

\begin{IEEEkeywords}
Semantička segmentacija, Oblak točaka, S3DIS
\end{IEEEkeywords}

\section{Uvod}\label{sec:introduction}
Semantička segmentacija 2D slika za cilj ima odrediti semantički razred za svaki piksel ulazne slike. Za razliku od slika gdje su pikseli raspoređeni u pravilnoj 2D rešetci, oblaci točaka zapravo su skupovi točaka ugrađeni u kontinuirani metrički prostor~\cite{zhao2021iccv}. Oblaci točaka zbog toga su strukturno različiti od slika, za koje postoji veliki broj standardnih metoda u području računalnog vida (npr.~primjena konvolucijskih slojeva). Rad s oblacima točaka, odnosno s 3D podacima općenito, pronalazi primjene u autonomnoj vožnji, robotici te proširenoj stvarnosti. 

Prema~\cite{zhao2021iccv}, pristupe učenja iz oblaka točaka dijelimo na: projekcijske metode, metode koje se temelje na postupku vokselizacije (engl.~\emph{voxel-based}) te metode koje rade izravno s oblacima točaka na ulazu. Projekcijske metode zasnivaju se na projekciji nepravilnih 3D oblaka točaka na pravilne 2D rešetke iz više različitih pogleda~\cite{Chen_2017_CVPR, Lang_2019_CVPR, Su_2015_ICCV}. Kod ovog pristupa može doći do gubitka informacije prilikom projekcije. Osim toga, performanse modela koji koriste projekcijski pristup mogu ovisiti o odabranom pogledu prema kojemu projiciramo oblak točaka. Za razliku od projekcijskih metoda, vokselizacijske metode nepravilni oblak točaka transformiraju u 3D rešetku, a potom nad 3D rešetkom provode 3D konvoluciju~\cite{Song_2017_CVPR, maturana2015iros}. Budući da postupkom vokselizacije nastaje 3D rešetka, vokselizacijske metode računalno su i memorijski dosta zahtjevne. Umjesto projekcije ili vokselizacije oblaka točaka, razvijene su i metode koje rade izravno s nepravilnim oblacima točaka na ulazu, a pionir među takvim arhitekturama je model \emph{PointNet}~\cite{qi2017cvpr}.

Cilj ovoga rada reprodukcija je rezultata više modela za semantičku segmentaciju oblaka točaka. Kratak pregled korištenih modela dan je u odjeljku~\ref{sec:method}. U istom odjeljku opisan je i korišteni skup podataka S3DIS. U odjeljku~\ref{sec:experiments} prikazani su i komentirani dobiveni rezultati te su navedene smjernice za budući rad.  
\section{Metoda}\label{sec:method}
Kao što je navedeno u odjeljku~\ref{sec:introduction}, postoji više načina na koje možemo učiti iz oblaka točaka. Govorimo li o izravnom učenju iz skupa točaka, začetnik i pokretač daljnjeg istraživanja svakako je model \emph{PointNet}~\cite{qi2017cvpr}. Osnovna ideja modela \emph{PointNet} ugraditi je točke u semantički bogatiji prostor značajki, a potom koristeći simetričnu funkciju sažimanja maksimumom (engl.~\emph{max pooling}) dobiti globalnu reprezentaciju ulaznog oblaka točaka. U članku~\cite{qi2017cvpr} navode se 3 svojstva koja trebaju zadovoljiti modeli koji na ulaz izravno dobivaju skup točaka:
\begin{enumerate}
    \item invarijantnost na poredak točaka na ulazu,
    \item invarijantnost na transformacije (npr.~rotacija ili translacija) te
    \item sposobnost modeliranja lokalne interakcije točaka.
\end{enumerate}
\begin{table*}[h!]
\caption{Statistika skupa podataka. Ukupan je broj točaka u skupu podataka 273.6M. Sve vrijednosti izražene su u postotcima (\%).}
\label{tab:dataset_statistics}
\centering
\scalebox{1}{
\begin{tabular}{l|r||r|r|r|r|r|r|r|r|r|r|r|r|r}
    \hline
   Područje & Udio točaka & ceiling & floor & wall & beam & column & window & door & table & chair & sofa & bookcase & board & clutter\\
\hline
Area-1 & 16.09 & 18.89 & 14.66 & 26.03 & 4.83 & 2.95 & 3.36 & 5.08 & 3.68 & 2.54 & 0.39 & 4.03 & 1.82 & 11.75 \\
Area-2 & 17.29 & 21.15 & 20.04 & 26.51 & 0.84 & 0.78 & 0.39 & 5.36 & 1.53 & 7.66 & 0.32 & 4.00 & 1.00 & 10.42 \\
Area-3 & 6.82 & 19.77 & 16.13 & 27.50 & 1.90 & 1.88 & 1.77 & 4.65 & 2.94 & 2.54 & 1.02 & 6.85 & 1.53 & 11.53 \\
Area-4 & 15.97 & 17.58 & 16.04 & 30.76 & 0.23 & 2.11 & 2.49 & 6.40 & 2.75 & 2.99 & 0.61 & 6.06 & 0.48 & 11.51 \\
Area-5 & 28.72 & 19.57 & 16.54 & 29.20 & 0.03 & 1.76 & 3.52 & 3.03 & 3.75 & 1.87 & 0.27 & 10.35 & 1.19 & 8.92 \\
Area-6 & 15.11 & 18.49 & 15.14 & 25.56 & 4.21 & 2.92 & 2.54 & 5.43 & 5.38 & 3.40 & 0.42 & 3.86 & 1.67 & 10.99 \\
\hline
Ukupno & 100.00 & 19.27 & 16.52 & 27.81 & 1.73 & 2.02 & 2.52 & 4.78 & 3.39 & 3.43 & 0.42 & 6.33 & 1.24 & 10.54 \\
\end{tabular}    
}
\end{table*}

U članku~\cite{qi2017cvpr} svojstvo 1) pokušava se zadovoljiti primjenom simetrične funkcije sažimanja maksimumom, svojstvo 2) primjenom plitke neuronske mreža koja uči transformaciju, dok se svojstvo 3) pokušava osigurati primjenom potpuno povezanih slojeva.
\subsection{PointNet++}\label{subsec:pointnet++}
Model \emph{PointNet++}~\cite{qi2017neurips} nadogradnja je modela \emph{PointNet} u vidu hijerarhijske obrade ulaznog oblaka točaka pri čemu se u obzir uzima udaljenost me{\dj}u točkama. Dok \emph{PointNet} sažima ulazni oblak točaka jednim izvršavanjem funkcije sažimanja maksimumom, model \emph{PointNet++} kroz više hijerarhijskih razina gradi reprezentaciju ulaznog oblaka točaka. Prema~\cite{qi2017neurips} jednu razinu hijerarhije nazivamo \emph{razinom apstrakcije skupa} (engl.~\emph{set abstraction level}). Svaka razina sastoji se od: 1) sloja uzorkovanja, 2) sloja grupiranja te 3) \emph{PointNet} sloja. Sloj uzorkovanja omogućava odabir točaka koje će predstavljati centroide lokalnih regija ulaznog skupa točaka, sloj grupiranja odabire grupe točaka koje, zajedno s centroidom, čine jednu regiju, dok \emph{PointNet} sloj kodira lokalne regije u vektore značajki.  
\subsection{PointCNN}\label{subsec:pointcnn}
Model \emph{PointNet} invarijantnost na transformacije postiže naučenom transformacijom točaka koja se na svaku točku primjenjuje zasebno, a invarijantnost na poredak postiže se korištenjem simetrične funkcije sažimanja maksimumom. S druge strane, model \emph{PointCNN}~\cite{yangyan2018nips} invarijantnost na poredak i transformacije pokušava postići učenjem $K$x$K$ matrice za podskup od $K$ susjednih točaka, a nakon toga nad skupom od tih $K$ točaka provodi se klasična konvolucija. Obični operator konvolucije zamjenjuje se tzv.~$\mathcal{X}$\emph{-Conv} operatorom.
\subsection{Cylinder3D}\label{subsec:cylinder3d}
Model \emph{Cylinder3D}~\cite{zhu2021cvpr} izvorno je namijenjen za semantičku segmentaciju oblaka točaka dobivenih rotirajućim radarskim uređajem~\englsk{Light Detection And Ranging}{LiDAR}. Glavna karakteristika takvih oblaka točaka veća je gustoća točaka koje su bliže radaru, te smanjenje gustoće točaka s porastom udaljenosti. Model \emph{Cylinder3D} ulazni oblak točaka transformira iz Kartezijevog u cilindrični koordinatni sustav te u cilindričnom koordinatnom sustavu provodi postupak vokselizacije. Potom se pomoću jednostavnog modela sličnog \emph{PointNet}-u dobivaju značajke za svaku točku, a nakon toga se koriste 3D rijetke konvolucije i arhitektura inspirirana poznatom arhitekturom za semantičku segmentaciju \emph{U-Net}~\cite{ronneberger2015miccai}.
\begin{table*}[h!]
\caption{Usrednjeni rezultati za svih 6 područja. Prva tri stupca predstavljaju ukupnu točnost ($OA$), srednju točnost na razini razreda ($mAcc$) te srednji omjer presjeka i unije ($mIoU$), dok preostali stupci predstavljaju omjer presjeka i unije za svaki pojedini razred ($IoU_i$). Prva 4 retka predstavljaju rezultate objavljene u referenciranim člancima, a ostali retci predstavljaju reprodukciju rezultata. Svi rezultati su u postotcima (\%).}
\label{tab:overall}
\centering
\scalebox{0.85}{
\begin{tabular}{l|r|r|r||r|r|r|r|r|r|r|r|r|r|r|r|r}
    \hline
    
   Model & $OA$ & $mAcc$ & $mIoU$ & ceiling & floor & wall & beam & column & window & door & table & chair & sofa & bookcase & board & clutter\\
\hline

PointCNN~\cite{yangyan2018nips} & 88.14 & 75.61 & 65.39 & 94.78 &97.30 &75.82 & 63.25 & 51.71 & 58.38 & 57.18 & 71.63 & 69.12 & 39.08 & 61.15 & 52.19 & 58.59 \\
PointNet++~\cite{ran2022cvpr} & 59.90 & 66.10 & 57.50 & -- &-- &-- & -- & -- & -- & -- & -- & -- & -- & -- & -- & -- \\
PointTransformer~\cite{zhao2021iccv} & 90.20 & 81.90 & 73.50 & -- &-- &-- & -- & -- & -- & -- & -- & -- & -- & -- & -- & -- \\
RepSurf~\cite{ran2022cvpr} & \textbf{90.80} & \textbf{82.60} & \textbf{74.30} & -- &-- &-- & -- & -- & -- & -- & -- & -- & -- & -- & -- & -- \\
\hline
Cylinder3D & 86.10 & 70.41 & 61.14 & 92.43 & 95.35 & 79.35 & 33.40 & 33.40 & 65.83 & 60.85 & 60.86 & 72.66 & 33.14 & 56.40 & 54.86 & 55.74 \\ 
\hline
PointCNN & 87.13 & 72.55 & 63.11 &93.75 &95.83 &81.76 &37.18 &41.59 &69.45 &56.05 &65.08 &67.71 &37.90 &60.11 & 53.85 & 59.50\\
PointNet++ & 89.15 & 80.08 & 69.70 & 94.02 & 97.31 & 83.30 & 42.51 & 49.39 & 57.09 & 76.60 & 69.55 & 77.19 & 68.77 & 62.64 & 64.47 & 63.27 \\
PointTransformer & \textbf{90.31} & \textbf{81.73} & \textbf{72.46} & 94.45 & 97.74 & 84.90 & 40.81 & 49.15 & 68.01 & 81.87 & 73.16 & 79.96 & 73.45 & 66.27 & 66.29 & 65.88 \\
RepSurf & 89.11 & 80.24 & 70.05 & 94.23 & 97.28 & 83.79 & 39.07 & 52.99 & 56.98 & 77.89 & 70.20 & 75.15 & 72.38 & 63.95 & 63.65 & 63.02 \\
\hline
\end{tabular}    
}
\end{table*}

\subsection{Point Transformer}\label{subsec:pointtrans}
Model \emph{Point Transformer}~\cite{zhao2021iccv} arhitektura je koja se temelji na arhitekturi Transformer~\cite{vaswani2017nips} koja je omogućila značajan napredak u području obrade prirodnog jezika, a sve češće se koristi i u području računalnoga vida.  Prema~\cite{zhao2021iccv}, mehanizam samopažnje (engl.~\emph{self-attention}), koji predstavlja srž arhitekture Transformer, zapravo je operator nad skupovima, što ga čini posebno pogodnim za primjenu nad oblacima točaka.
\subsection{RepSurf}\label{subsec:repsurf}
Osnovna ideja modela \emph{RepSurf}~\cite{ran2022cvpr} (engl. representative surface) omogućiti je modelu uvid u lokalnu strukturu oblaka točaka tako da na ulaz modela, umjesto samih koordinata, dovodimo i informaciju o normali u svakoj pojedinoj točki, kao i direktne informacije o lokalnom susjedstvu pojedine točke modeliranjem površina trokuta (\emph{triangular surfaces}) ili površina kišobrana (\emph{umbrella surfaces}). Model \emph{RepSurf} kao osnovicu koristi arhitekturu \emph{PointNet++}, pri čemu se na svaki skup apstrakcije modela \emph{PointNet++} na ulaz dovode i značajke koje predstavljaju lokalnu strukturu susjedstva. U okviru ovoga rada koristi se model koji na ulazu dobiva značajke koje modeliraju površinu kišobrana, odnosno model \emph{RepSurf-U} iz članka~\cite{ran2022cvpr}.
\subsection{Skup podataka}\label{sec:dataset}
Predstavljene modele vrjednovali smo pomoću skupa podataka za semanti\v{c}ku segmentaciju zvanog S3DIS~\cite{armeni2016cvpr}. Skup podataka S3DIS skup je uniformno uzorkovanih oblaka to\v{c}aka, a nastao je 3D skeniranjem interijera prostorija pomoću \emph{Matterport}\footnote{https://matterport.com/pro2} kamere. Skup podataka sastoji se od 271 prostorije iz 6 razli\v{c}itih podru\v{c}ja, pri \v{c}emu svako podru\v{c}je predstavlja jednu etažu unutar ukupno 3 ustanove. Ukupan broj to\v{c}aka u skupu približno je 274M, pri \v{c}emu svaka to\v{c}ka pripada jednom od 13 semanti\v{c}kih razreda (npr.~\emph{clutter}, \emph{ceiling}, \emph{bookcase}). Osim koordinata točke, u skupu podataka za svaku točku zabilježena je i boja (RGB komponente). Tablica~\ref{tab:dataset_statistics} prikazuje udio točaka za svako područje te za svaki semantički razred, kao i udio točaka u pojedinom području s obzirom na ukupan broj točaka. Možemo uočiti kako je skup podataka izrazito neuravnotežen, s obzirom na to da točke s oznakama samo tri semantička razreda (\emph{ceiling}, \emph{floor} i \emph{wall}) čine 63.6\% ukupnog broja točaka.

Naučene modele vrjednovali smo šesterostrukom unakrsnom provjerom (engl.~\emph{6-fold cross-validation}), tako da smo svaki model učili 6 puta, pri čemu smo 5 područja odabrali za skup za učenje, a jedno preostalo područje predstavljalo je skup za provjeru. Osim šesterostruke unakrsne provjere, vrednovanje modela na skupu podataka S3DIS često se provodi tako da se područje 5 izdvoji kao ispitni skup, s obzirom na to da je područje 5 najveće područje u skupu podataka. U odjeljku~\ref{sec:appendix} prikazani su rezultati za svako područje i za svaki model, pa tako i za područje 5. 
\subsection{Predobrada skupa podataka}
Ovisno o korištenom modelu, predobradu skupa podataka proveli smo na dva načina. Za modele \emph{PointNet++}, \emph{PointTransformer} te \emph{RepSurf} predobradu skupa podataka proveli smo po uzoru na~\cite{ran2022cvpr}. Svaki oblak točaka predstavljen je 3D rešetkom postupkom vokselizacije, pri čemu je veličina voksela iznosila 0.04m. Iz dobivene rešetke potom odabiremo podskup od najviše 80000 točaka te koristimo model kako bismo dobili oznake semantičkih razreda za svaku točku, a postupak ponavljamo dok svaka točka ne bude odabrana. Za točke koje su odabrane više puta konačna je oznaka ona koju je ta točka poprimila najviše puta. S druge strane, za modele \emph{PointCNN} i \emph{Cylinder3D} predobradu skupa podataka proveli smo po uzoru na~\cite{yangyan2018nips}. Svaka prostorija dijeli se na blokove od 1.5m po svakoj dimenziji, s nadopunom od 0.3m sa svake strane. Nadopunjene točke predstavljaju kontekst za središnji blok te se one ne uzimaju u obzir pri računanju gubitka za vrijeme učenja niti za dobivanje oznaka pri zaključivanju. Tako dobiveni blokovi transformirani su u lokalne koordinatne sustave, pri čemu je ishodište koordinatnog sustava središte bloka.
\section{Eksperimenti}\label{sec:experiments}
Za vrjednovanje semantičke segmentacije oblaka točaka koristili smo 3 standardne metrike:
\begin{itemize}
    \item[$\bullet$] ukupna točnost~\englsk{Overall Accuracy}{OA} - omjer točno označenih i svih točaka;
    \item[$\bullet$] srednja točnost na razini razreda~\englsk{mean class Accuracy}{mAcc} - srednja točnost za sve razrede, pri čemu točnost na razini jednog razreda predstavlja udio točno predviđenih točaka u svim točkama koje pripadaju tom razredu;
    \item[$\bullet$] omjer presjeka i unije na razini razreda~\englsk{Intersection over Union}{IoU} - presjek točaka za koje je naš model predvidio pripadnost određenom razredu i svih točaka koje pripadaju tom razredu podijeljen njihovom unijom. 
\end{itemize}
U Tablici~\ref{tab:overall} prikazani su usrednjeni rezultati šesterostruke unakrsne validacije za svaki od modela. Dobiveni rezultati svakog od modela za svako područje dostupni su u Dodatku~\ref{sec:appendix}. Osim toga, modele smo usporedili i s obzirom na vrijeme trajanja unaprijednog prolaza, a rezultate tih mjerenja prikazuje  Tablica~\ref{tab:speed}.

\subsection{Hiperparametri}
Za model \emph{PointCNN} svi hiperparametri preuzeti su sa službenog repozitorija\footnote{https://github.com/yangyanli/PointCNN} objavljenog u~\cite{qi2017cvpr}. Učenje smo proveli na jednoj grafičkoj kartici NVIDIA Titan Xp. Za modele \emph{PointNet++}, \emph{Point Transformer} i \emph{RepSurf} hiperparametri su preuzeti sa repozitorija\footnote{https://github.com/hancyran/RepSurf} objavljenog u~\cite{ran2022cvpr}. Učenje navedenih modela proveli smo na dvije NVIDIA RTX A4500 grafičke kartice. Za razliku od~\cite{ran2022cvpr}, veličinu grupe (engl.~\emph{batch size}) pri učenju modela \emph{Point Transformer} smanjili smo s 8 na 4 zbog nedostatka memorijskih resursa. Što se tiče modela \emph{Cylinder3D}, hiperparametri navedeni u~\cite{zhu2021cvpr} korišteni su za učenje modela na skupu podataka SemanticKITTI\footnote{http://www.semantic-kitti.org/}. Kako bismo prilagodili te hiperparametre za skup podataka S3DIS isprobali smo nekoliko različitih vrijednosti hiperparametara (dimenzije okvira pri postupku vokselizacije, stopa učenja i veličina grupe), a najbolji rezultati dobiveni su uz vrijednost stope učenja od $0.001$ i veličinu grupe $32$, dok su dimenzije okvira pri postupku vokselizacije postavljene na vrijednost \texttt{[128, 100, 16]}. Učenje modela provedeno je kroz 20 epoha na jednoj grafičkoj kartici NVIDIA RTX A4500.
\subsection{Diskusija}
\begin{table*}[h!]
\caption{Trajanje unaprijednog prolaza za svaki od naučenih modela i 10 različitih prostorija iz područja 1. Prikazana je srednja vrijednost i standardna devijacija kroz 3 pokretanja. Sva mjerenja izražena su u sekundama.}
\label{tab:speed}
\centering
\scalebox{0.9}{
\begin{tabular}{l|c|c|c|c|c|c}
Prostorija & Broj to\v{c}aka & \emph{PointCNN} (s) & \emph{Cylinder3D} (s) & \emph{PointNet++} (s) & \emph{Point Transformer} (s) & \emph{RepSurf} (s) \\
\hline
\hline

hallway1 &344k & 27.36 $\pm$ 0.177 & 3.226$\pm$0.026 & \textbf{2.360}$\mathbf{\pm}$\textbf{0.013} & 7.466$\pm$0.094 & 3.855$\pm$0.007 \\
hallway4 &367k & 28.159$\pm$0.137 & 3.537$\pm$0.054 & \textbf{2.126}$\mathbf{\pm}$\textbf{0.007} & 6.278$\pm$0.064 & 3.548$\pm$0.026 \\
hallway3 &369k & 29.402$\pm$0.091 & 3.536$\pm$0.025 & \textbf{2.652}$\mathbf{\pm}$\textbf{0.040} & 8.928$\pm$0.048 & 4.371$\pm$0.028 \\
hallway5 &465k & 35.652$\pm$0.085 & \textbf{4.419}$\mathbf{\pm}$\textbf{0.021} & 5.074$\pm$0.054 & 15.940$\pm$0.033 & 8.962$\pm$0.046 \\
copy & 510k & 40.213$\pm$0.157 & \textbf{5.378}$\mathbf{\pm}$\textbf{0.042} & 6.170$\pm$0.023 & 18.649$\pm$0.062 & 10.687$\pm$0.011 \\
hallway2 &579k & 43.789$\pm$0.274 & \textbf{5.208}$\mathbf{\pm}$\textbf{0.038} & 6.416$\pm$0.031 & 21.306$\pm$0.135 & 11.635$\pm$0.009 \\
WC1 & 1.11M & 86.613$\pm$0.179 & \textbf{10.649}$\mathbf{\pm}$\textbf{0.073} & 15.847$\pm$0.216 & 51.140$\pm$0.061 & 31.529$\pm$0.167 \\
cfRoom1 &1.14M & 88.896$\pm$0.224 & \textbf{12.251}$\mathbf{\pm}$\textbf{1.557} & 12.624$\pm$0.030 & 42.196$\pm$0.046 & 26.411$\pm$0.038 \\
cfRoom2 &1.54M & 121.504$\pm$0.094 & \textbf{15.387}$\mathbf{\pm}$\textbf{0.162} & 57.267$\pm$0.969 & 361.840$\pm$6.348 & 118.262$\pm$3.881 \\
hallway6 &3.84M & 298.632$\pm$0.370 & \textbf{38.765}$\mathbf{\pm}$\textbf{0.386} & 113.068$\pm$1.889 & 731.968$\pm$13.751 & 239.419$\pm$5.431 \\
\hline
\end{tabular}    
}
\end{table*}

Usrednjeni rezultati šesterostruke unakrsne validacije za svaki od modela prikazani su u Tablici~\ref{tab:overall}. Usredotočimo li se na metriku \emph{mIoU}, najbolje rezultate ostvarili smo modelom \emph{Point Transformer}. Reproducirani rezultat za model \emph{Point Transformer} lošiji je od rezultata objavljenoga u~\cite{zhao2021iccv} za 1.04\%. Postoji mogućnost da je ta razlika nastala zbog smanjene veličine grupe.  Nadalje, možemo vidjeti kako postoji značajna razlika između rezultata za model \emph{RepSurf} koji su objavljeni u~\cite{ran2022cvpr} i reproduciranih rezultata. Ipak, rezultati za model \emph{RepSurf} dobiveni za područje 5 podudaraju se sa službenim rezultatima objavljenima u~\cite{ran2022cvpr}. Iz tog razloga, postoji mogućnost da je razlika dobivena za šesterostruku unakrsnu validaciju nastala zbog različitih hiperparametara koje su koristili autori članka~\cite{ran2022cvpr} za učenje modela na ostalim područjima. Vrijedi istaknuti i značajnu razliku između rezultata dobivenih za model \emph{PointNet++}. Autori članka~\cite{ran2022cvpr} nisu naveli na koji način su došli do rezultata za model~\emph{PointNet++}, a rezultati za taj model koji su objavljeni na službenom repozitoriju razlikuju se od onih koji su objavljeni u~\cite{ran2022cvpr}. Dobivene razlike u rezultatima otvaraju zanimljive smjerove budućega istraživanja, kao što je npr.~provjera doprinosa modeliranja lokalnog susjedstva za model \emph{RepSurf}. Gledajući općenito rezultate u Tablici~\ref{tab:overall}, možemo zaključiti kako svi modeli postižu najbolje rezultate na najzastupljenijim razredima u skupu podataka (\emph{ceiling}, \emph{floor} i \emph{wall}). Vizualizaciju rezultata semantičke segmentacije za dvije prostorije te za svaki od modela prikazali smo u Dodatku~\ref{sec:appendix}. 

Promotrimo li vremena trajanja unaprijednog prolaza u Tablici~\ref{tab:speed}, vidimo kako je \emph{Cylinder3D} za prostorije s više od 0.5M točaka konzistentno najbrži model. Brzina zaključivanja kod ostalih modela značajnije ovisi o veličini ulaznog oblaka točaka. Takva ovisnost prisutna je zbog korištenja vokselizacije u svrhu odabira podskupa točaka koje u jednom koraku dovodimo na ulaz modela (\emph{grid sampling})~\cite{ran2022cvpr}. Zbog značajne razlike u brzinama za velike oblake točaka, model \emph{Cylinder3D}, unatoč najlošijim performansama u vidu evaluacijskih metrika (Tablica~\ref{tab:overall}), ostaje model vrijedan budućeg istraživanja i potencijalnog poboljšanja s pronalaskom optimalnih hiperparametara. 
\section{Zaklju\v{c}ak}\label{sec:conclusions}
U okviru ovoga rada proveden je niz eksperimenata u svrhu reprodukcije rezultata različitih modela za semantičku segmentaciju oblaka točaka na skupu podataka S3DIS. Dobiveni rezultati pokazuju kako je najuspješniji model u vidu evaluacijskih metrika \emph{Point Transformer}, dok je model s najkraćim trajanjem zaključivanja \emph{Cylinder3D}. Osim toga, postoje značajne razlike u rezultatima za modele \emph{RepSurf} i \emph{PointNet++} od rezultata navedenih u odgovarajućim člancima, a objasniti dobivene razlike dio je budućega rada. S obzirom na brzinu zaključivanja modela \emph{Cylinder3D}, budući rad uključuje optimizaciju hiperparametara za taj model. Nadalje, potrebno je istražiti mogućnosti ubrzanja modela \emph{Point Transformer} i \emph{RepSurf}, odnosno ispitati postoji li značajan pad performansi ako ne koristimo \emph{grid sampling} prilikom pripreme podataka.
\bibliography{bib/bibliography}

\begin{thebibliography}{10}
\providecommand{\url}[1]{#1}
\csname url@samestyle\endcsname
\providecommand{\newblock}{\relax}
\providecommand{\bibinfo}[2]{#2}
\providecommand{\BIBentrySTDinterwordspacing}{\spaceskip=0pt\relax}
\providecommand{\BIBentryALTinterwordstretchfactor}{4}
\providecommand{\BIBentryALTinterwordspacing}{\spaceskip=\fontdimen2\font plus
\BIBentryALTinterwordstretchfactor\fontdimen3\font minus
  \fontdimen4\font\relax}
\providecommand{\BIBforeignlanguage}[2]{{%
\expandafter\ifx\csname l@#1\endcsname\relax
\typeout{** WARNING: IEEEtran.bst: No hyphenation pattern has been}%
\typeout{** loaded for the language `#1'. Using the pattern for}%
\typeout{** the default language instead.}%
\else
\language=\csname l@#1\endcsname
\fi
#2}}
\providecommand{\BIBdecl}{\relax}
\BIBdecl

\bibitem{zhao2021iccv}
H.~Zhao, L.~Jiang, J.~Jia, P.~H. Torr, and V.~Koltun, ``Point transformer,'' in
  \emph{Proceedings of the IEEE/CVF International Conference on Computer Vision
  (ICCV)}, October 2021, pp. 16\,259--16\,268.

\bibitem{Chen_2017_CVPR}
X.~Chen, H.~Ma, J.~Wan, B.~Li, and T.~Xia, ``Multi-view 3d object detection
  network for autonomous driving,'' in \emph{Proceedings of the IEEE Conference
  on Computer Vision and Pattern Recognition (CVPR)}, July 2017.

\bibitem{Lang_2019_CVPR}
A.~H. Lang, S.~Vora, H.~Caesar, L.~Zhou, J.~Yang, and O.~Beijbom,
  ``Pointpillars: Fast encoders for object detection from point clouds,'' in
  \emph{Proceedings of the IEEE/CVF Conference on Computer Vision and Pattern
  Recognition (CVPR)}, June 2019.

\bibitem{Su_2015_ICCV}
H.~Su, S.~Maji, E.~Kalogerakis, and E.~Learned-Miller, ``Multi-view
  convolutional neural networks for 3d shape recognition,'' in
  \emph{Proceedings of the IEEE International Conference on Computer Vision
  (ICCV)}, December 2015.

\bibitem{Song_2017_CVPR}
S.~Song, F.~Yu, A.~Zeng, A.~X. Chang, M.~Savva, and T.~Funkhouser, ``Semantic
  scene completion from a single depth image,'' in \emph{Proceedings of the
  IEEE Conference on Computer Vision and Pattern Recognition (CVPR)}, July
  2017.

\bibitem{maturana2015iros}
D.~Maturana and S.~Scherer, ``Voxnet: A 3d convolutional neural network for
  real-time object recognition,'' in \emph{2015 IEEE/RSJ International
  Conference on Intelligent Robots and Systems (IROS)}, 2015, pp. 922--928.

\bibitem{qi2017cvpr}
C.~R. Qi, H.~Su, K.~Mo, and L.~J. Guibas, ``Pointnet: Deep learning on point
  sets for 3d classification and segmentation,'' in \emph{Proceedings of the
  IEEE Conference on Computer Vision and Pattern Recognition (CVPR)}, July
  2017.

\bibitem{qi2017neurips}
C.~R. Qi, L.~Yi, H.~Su, and L.~J. Guibas, ``Pointnet++: Deep hierarchical
  feature learning on point sets in a metric space,'' in \emph{Advances in
  Neural Information Processing Systems}, I.~Guyon, U.~V. Luxburg, S.~Bengio,
  H.~Wallach, R.~Fergus, S.~Vishwanathan, and R.~Garnett, Eds., vol.~30.\hskip
  1em plus 0.5em minus 0.4em\relax Curran Associates, Inc., 2017.

\bibitem{yangyan2018nips}
Y.~Li, R.~Bu, M.~Sun, W.~Wu, X.~Di, and B.~Chen, ``Pointcnn: Convolution on
  x-transformed points,'' in \emph{Advances in Neural Information Processing
  Systems}, S.~Bengio, H.~Wallach, H.~Larochelle, K.~Grauman, N.~Cesa-Bianchi,
  and R.~Garnett, Eds., vol.~31.\hskip 1em plus 0.5em minus 0.4em\relax Curran
  Associates, Inc., 2018.

\bibitem{zhu2021cvpr}
X.~Zhu, H.~Zhou, T.~Wang, F.~Hong, Y.~Ma, W.~Li, H.~Li, and D.~Lin,
  ``Cylindrical and asymmetrical 3d convolution networks for lidar
  segmentation,'' in \emph{Proceedings of the IEEE/CVF Conference on Computer
  Vision and Pattern Recognition (CVPR)}, June 2021, pp. 9939--9948.

\bibitem{ronneberger2015miccai}
O.~Ronneberger, P.~Fischer, and T.~Brox, ``U-net: Convolutional networks for
  biomedical image segmentation,'' in \emph{Medical Image Computing and
  Computer-Assisted Intervention -- MICCAI 2015}, N.~Navab, J.~Hornegger, W.~M.
  Wells, and A.~F. Frangi, Eds.\hskip 1em plus 0.5em minus 0.4em\relax Cham:
  Springer International Publishing, 2015, pp. 234--241.

\bibitem{ran2022cvpr}
H.~Ran, J.~Liu, and C.~Wang, ``Surface representation for point clouds,'' in
  \emph{Proceedings of the IEEE/CVF Conference on Computer Vision and Pattern
  Recognition (CVPR)}, June 2022, pp. 18\,942--18\,952.

\bibitem{vaswani2017nips}
A.~Vaswani, N.~Shazeer, N.~Parmar, J.~Uszkoreit, L.~Jones, A.~N. Gomez, L.~u.
  Kaiser, and I.~Polosukhin, ``Attention is all you need,'' in \emph{Advances
  in Neural Information Processing Systems}, I.~Guyon, U.~V. Luxburg,
  S.~Bengio, H.~Wallach, R.~Fergus, S.~Vishwanathan, and R.~Garnett, Eds.,
  vol.~30.\hskip 1em plus 0.5em minus 0.4em\relax Curran Associates, Inc.,
  2017.

\bibitem{armeni2016cvpr}
I.~Armeni, O.~Sener, A.~R. Zamir, H.~Jiang, I.~Brilakis, M.~Fischer, and
  S.~Savarese, ``3d semantic parsing of large-scale indoor spaces,'' in
  \emph{Proceedings of the IEEE Conference on Computer Vision and Pattern
  Recognition (CVPR)}, June 2016.

\end{thebibliography}
\bibliographystyle{bib/IEEEtran}
\section{Dodatak}\label{sec:appendix}
\begin{table*}[h!]
\caption{Dobiveni rezultati za model PointCNN. Prva tri stupca predstavljaju ukupnu točnost ($OA$), srednju točnost na razini razreda ($mAcc$) te srednji omjer presjeka i unije ($mIoU$), dok preostali stupci predstavljaju omjer presjeka i unije za svaki pojedini razred ($IoU_i$). Svaki redak predstavlja područje koje se koristilo kao ispitni skup. Rezultati su prikazani u postotcima (\%).} \label{tab:pointcnn}
\centering
\scalebox{0.90}{
\begin{tabular}{l|r|r|r||r|r|r|r|r|r|r|r|r|r|r|r|r}
\hline
Područje & $OA$ & $mAcc$ & $mIoU$ & ceiling & floor & wall & beam & column & window & door & table & chair & sofa & bookcase & board & clutter\\ \hline
Area-1 & 89.03 & 80.01 & 71.25 &95.59&95.58&81.77&57.44&51.90&88.08&78.59&69.71&75.16&43.23&59.63&62.44& 67.21\\
Area-2 & 80.36& 58.41 & 45.25 &88.64&87.31&82.00&14.52&40.30&66.83&27.42&31.55&26.04&29.48&44.36&11.13& 38.69\\ 
Area-3 & 90.31 & 82.56 & 72.85 &95.24&98.57&82.19&70.58&35.34&88.41&69.32&75.40&73.48&40.15&70.77&76.29& 71.42\\ 
Area-4 & 84.82 & 65.00 & 54.82 &93.76&97.62&80.19&2.70&42.15&57.02&29.57&62.53&63.81&30.76&44.17&49.54& 58.82\\ 
Area-5 & 86.37& 62.41 & 55.92 &92.84&98.12&78.79&0.01&21.85&25.78&52.34&76.47&84.36&23.93&67.81&53.36& 51.32\\ 
Area-6 & 91.90 & 86.91 & 78.57 &96.48&97.80&85.60&77.82&58.00&90.59&79.07&79.14&83.38&59.83&73.93&70.36& 69.53\\ 
\hline
Prosjek& 87.13 & 72.55 & 63.11 &93.75 &95.83 &81.76 &37.18 &41.59 &69.45 &56.05 &65.08 &67.71 &37.90 &60.11 & 53.85 & 59.50 \\
\hline
\end{tabular}    
}
\end{table*} 

\begin{table*}[h!]
\caption{Dobiveni rezultati za model Cylinder3D. Prva tri stupca predstavljaju ukupnu točnost ($OA$), srednju točnost na razini razreda ($mAcc$) te srednji omjer presjeka i unije ($mIoU$), dok preostali stupci predstavljaju omjer presjeka i unije za svaki pojedini razred ($IoU_i$). Svaki redak predstavlja područje koje se koristilo kao ispitni skup. Rezultati su prikazani u postotcima (\%).}
\label{tab:cylinder}
\centering
\scalebox{0.9}{
\begin{tabular}{l|r|r|r||r|r|r|r|r|r|r|r|r|r|r|r|r}
\hline
Područje & $OA$ & $mAcc$ & $mIoU$ & clutter & ceiling & floor & wall & beam & column & window & door & table & chair & sofa & bookcase & board\\ \hline
Area-1 & 86.28& 73.00 & 64.51 & 63.18&94.83&94.73&77.79&37.74&59.92&75.84&74.05&60.80&68.93&28.04&54.14&48.63\\
Area-2 & 81.82 & 58.32 & 45.22 &43.05&88.34&90.70&75.62&19.99&20.45&41.10&35.54&41.39&72.14&8.13&40.31&11.14\\
Area-3 & 89.73 & 81.83 & 73.15 &71.09&94.73&97.40&81.52&64.59&17.47&77.23&88.71&72.96&74.24&60.59&68.87&81.54\\
Area-4 & 83.61 & 63.95 & 54.69 &48.99&90.56&96.14&77.52&0.04&34.21&59.47&28.09&56.95&66.42&43.20&49.99&59.41\\
Area-5 & 84.61 & 61.96 & 53.65 &43.93&91.10&96.76&78.73&0.11&11.66&61.33&52.23&61.54&74.27&17.37&57.09&51.31\\
Area-6 & 90.42 & 83.37 & 75.64 &64.22&95.01&96.37&84.94&77.93&60.18&80.03&86.49&71.54&79.93&41.51&67.99&77.14 \\
\hline
Prosjek&  86.10 & 70.41 & 61.14 & 55.74 & 92.43 & 95.35 & 79.35 & 33.40 & 33.40 & 65.83 & 60.85 & 60.86 & 72.66 & 33.14 & 56.40 & 54.86 \\ 
\hline
\end{tabular}    
}
\end{table*}

\begin{table*}[h!]
\caption{Dobiveni rezultati za model PointNet++. Prva tri stupca predstavljaju ukupnu točnost ($OA$), srednju točnost na razini razreda ($mAcc$) te srednji omjer presjeka i unije ($mIoU$), dok preostali stupci predstavljaju omjer presjeka i unije za svaki pojedini razred ($IoU_i$). Svaki redak predstavlja područje koje se koristilo kao ispitni skup. Rezultati su prikazani u postotcima (\%).} \label{tab:pointcnn}
\centering
\scalebox{0.90}{
\begin{tabular}{l|r|r|r||r|r|r|r|r|r|r|r|r|r|r|r|r}
\hline
Područje & $OA$ & $mAcc$ & $mIoU$ & ceiling & floor & wall & beam & column & window & door & table & chair & sofa & bookcase & board & clutter \\ \hline
Area-1 & 89.93 & 84.75 & 75.11 & 96.60 & 95.75 & 82.51 & 65.71 & 60.63 & 76.97 & 86.98 & 70.83 & 77.07 & 66.89 & 57.56 & 69.18 & 69.74 \\
Area-2 & 85.82 & 70.51 & 56.75 & 89.1 & 95.83 & 82.15 & 17.98 & 45.97 & 23.20 & 63.07 & 53.16 & 64.89 & 65.73 & 51.14 & 35.05 & 50.49 \\
Area-3 & 91.51 & 86.87 & 77.95 & 95.62 & 98.29 & 83.14 & 71.08 & 34.73 & 68.47 & 89.45 & 75.52 & 82.67 & 77.08 & 73.48 & 88.66 & 75.19 \\
Area-4 & 86.71 & 74.93 & 62.16 & 95.08 & 98.00 & 81.98 & 24.90 & 56.95 & 35.12 & 61.55 & 61.27 & 69.1 & 64.69 & 53.34 & 45.59 & 60.49 \\
Area-5 & 88.70 & 72.39 & 65.07 & 91.81 & 98.38 & 83.64 & 0.00 & 20.79 & 62.0 & 68.98 & 79.45 & 86.22 & 65.99 & 70.10 & 66.22 & 52.29 \\
Area-6 & 92.24 & 91.01 & 81.17 & 95.92 & 97.58 & 86.38 & 75.39 & 77.29 & 76.76 & 89.56 & 77.04 & 83.21 & 72.24 & 70.21 & 82.13 & 71.45 \\
\hline
Prosjek & 89.15 & 80.08 & 69.70 & 94.02 & 97.31 & 83.30 & 42.51 & 49.39 & 57.09 & 76.60 & 69.55 & 77.19 & 68.77 & 62.64 & 64.47 & 63.27 \\
\hline
\end{tabular}    
}
\end{table*} 

\begin{table*}[h!]
\caption{Dobiveni rezultati za model Point Transformer. Prva tri stupca predstavljaju ukupnu točnost ($OA$), srednju točnost na razini razreda ($mAcc$) te srednji omjer presjeka i unije ($mIoU$), dok preostali stupci predstavljaju omjer presjeka i unije za svaki pojedini razred ($IoU_i$). Svaki redak predstavlja područje koje se koristilo kao ispitni skup. Rezultati su prikazani u postotcima (\%).} \label{tab:pointcnn}
\centering
\scalebox{0.90}{
\begin{tabular}{l|r|r|r||r|r|r|r|r|r|r|r|r|r|r|r|r}
\hline
Područje & $OA$ & $mAcc$ & $mIoU$ & ceiling & floor & wall & beam & column & window & door & table & chair & sofa & bookcase & board & clutter \\ \hline
Area-1 & 92.13 & 89.96 & 81.16 & 96.14 & 96.46 & 87.38 & 69.38 & 69.54 & 84.58 & 89.54 & 74.51 & 82.30 & 82.89 & 67.46 & 81.04 & 73.85 \\
Area-2 & 85.92 & 68.83 & 56.48 & 89.46 & 96.19 & 80.88 & 13.53 & 25.11 & 48.74 & 74.50 & 56.00 & 61.03 & 64.87 & 52.7 & 20.81 & 50.46 \\
Area-3 & 92.21 & 87.91 & 80.16 & 95.96 & 98.74 & 84.05 & 77.71 & 30.76 & 82.20 & 93.19 & 80.32 & 85.64 & 82.49 & 74.30 & 80.69 & 76.09 \\
Area-4 & 87.46 & 74.03 & 62.18 & 95.47 & 98.32 & 82.51 & 6.67 & 53.31 & 42.80 & 68.09 & 66.55 & 69.98 & 57.87 & 55.40 & 50.08 & 61.26 \\
Area-5 & 90.46 & 76.93 & 70.06 & 93.06 & 98.4 & 85.84 & 0.00 & 37.39 & 63.01 & 74.93 & 82.36 & 91.17 & 77.78 & 73.99 & 75.29 & 57.53 \\
Area-6 & 93.68 & 92.73 & 84.70 & 96.62 & 98.34 & 88.77 & 77.57 & 78.81 & 86.73 & 90.96 & 79.21 & 89.63 & 74.81 & 73.74 & 89.84 & 76.11 \\
\hline
Prosjek & 90.31 & 81.73 & 72.46 & 94.45 & 97.74 & 84.90 & 40.81 & 49.15 & 68.01 & 81.87 & 73.16 & 79.96 & 73.45 & 66.27 & 66.29 & 65.88 \\
\hline
\end{tabular}    
}
\end{table*} 

\begin{table*}[h!]
\caption{Dobiveni rezultati za model RepSurf. Prva tri stupca predstavljaju ukupnu točnost ($OA$), srednju točnost na razini razreda ($mAcc$) te srednji omjer presjeka i unije ($mIoU$), dok preostali stupci predstavljaju omjer presjeka i unije za svaki pojedini razred ($IoU_i$). Svaki redak predstavlja područje koje se koristilo kao ispitni skup. Rezultati su prikazani u postotcima (\%).} \label{tab:pointcnn}
\centering
\scalebox{0.90}{
\begin{tabular}{l|r|r|r||r|r|r|r|r|r|r|r|r|r|r|r|r}
\hline
Područje & $OA$ & $mAcc$ & $mIoU$ & ceiling & floor & wall & beam & column & window & door & table & chair & sofa & bookcase & board & clutter \\ \hline
Area-1 & 89.72 & 84.89 & 75.48 & 96.44 & 95.78 & 81.74 & 54.87 & 62.08 & 76.47 & 86.87 & 71.81 & 81.64 & 72.79 & 61.28 & 69.74 & 69.69 \\
Area-2 & 83.44 & 68.22 & 54.40 & 89.31 & 95.69 & 81.21 & 21.79 & 38.86 & 21.07 & 66.12 & 53.61 & 34.88 & 78.79 & 51.99 & 29.47 & 44.40 \\
Area-3 & 91.27 & 87.00 & 77.59 & 95.97 & 98.36 & 83.57 & 67.40 & 46.47 & 57.15 & 86.67 & 73.83 & 84.68 & 82.62 & 69.75 & 89.42 & 72.81 \\
Area-4 & 87.63 & 74.12 & 62.27 & 93.48 & 97.84 & 83.17 & 13.26 & 54.81 & 43.63 & 66.01 & 65.32 & 75.10 & 52.80 & 56.67 & 45.29 & 62.14 \\
Area-5 & 89.63 & 75.46 & 68.00 & 93.77 & 98.31 & 85.14 & 0.00 & 34.99 & 60.09 & 72.65 & 78.66 & 89.35 & 77.07 & 70.28 & 67.34 & 56.31 \\
Area-6 & 92.95 & 91.73 & 82.53 & 96.41 & 97.69 & 87.92 & 77.10 & 80.74 & 83.49 & 89.04 & 77.95 & 85.24 & 70.21 & 73.75 & 80.61 & 72.77 \\
\hline
Prosjek & 89.11 & 80.24 & 70.05 & 94.23 & 97.28 & 83.79 & 39.07 & 52.99 & 56.98 & 77.89 & 70.20 & 75.15 & 72.38 & 63.95 & 63.65 & 63.02 \\
\hline
\end{tabular}    
}
\end{table*}

\begin{figure*}[h!]
\centering
\begin{tabular}{ll}
\includegraphics[width=0.45\linewidth]{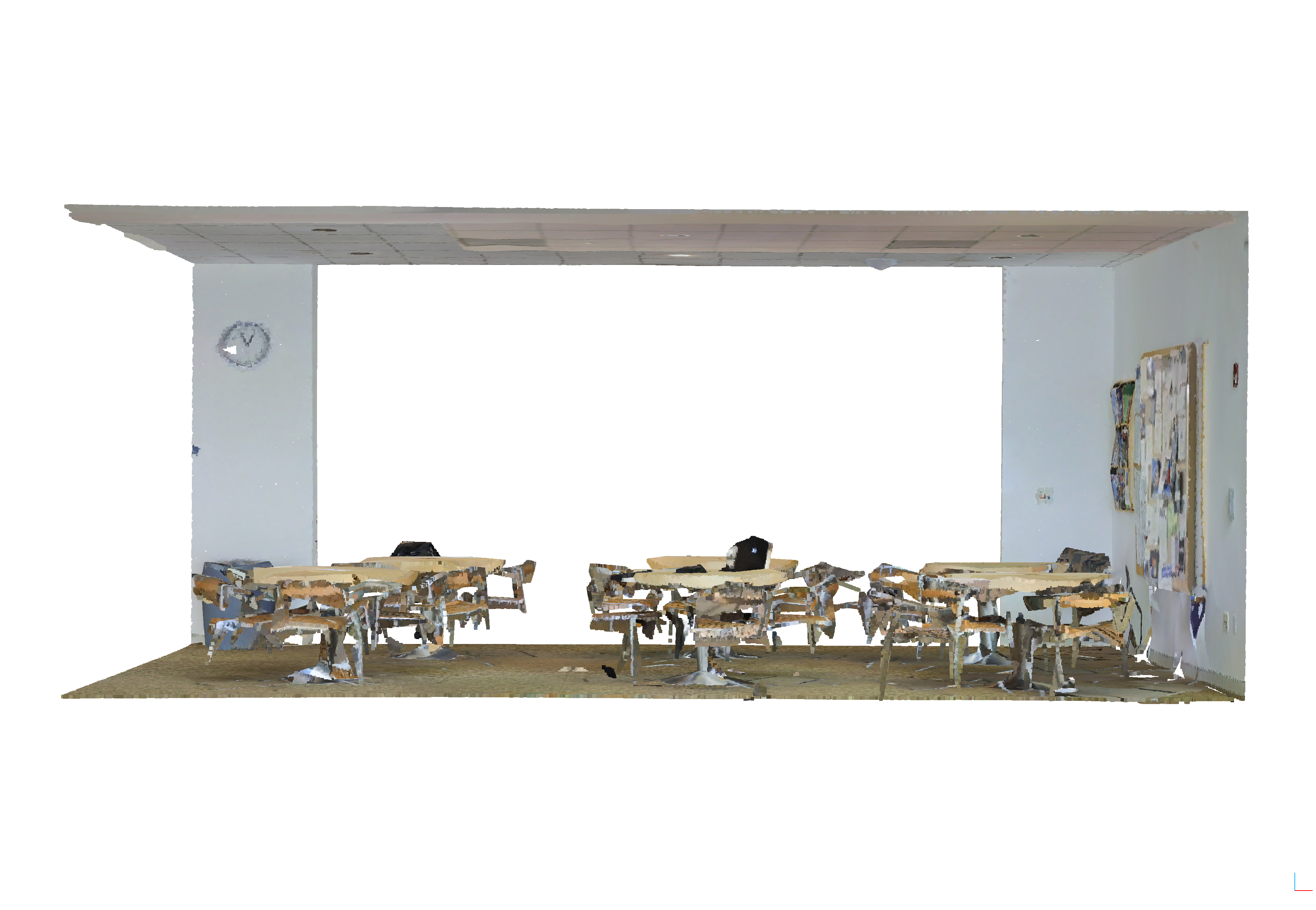} &
\includegraphics[width=0.45\linewidth]{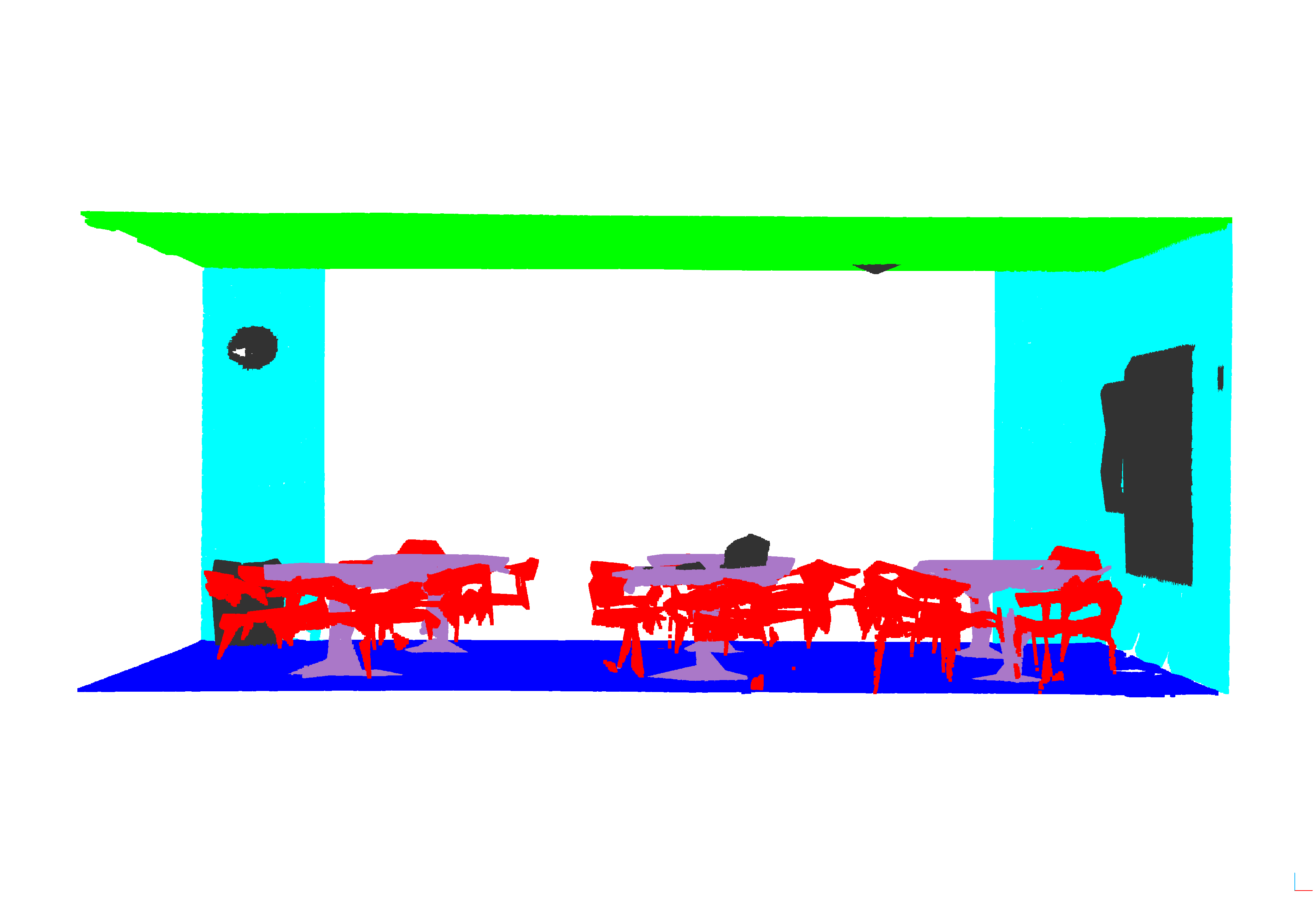} \\
\includegraphics[width=0.45\linewidth]{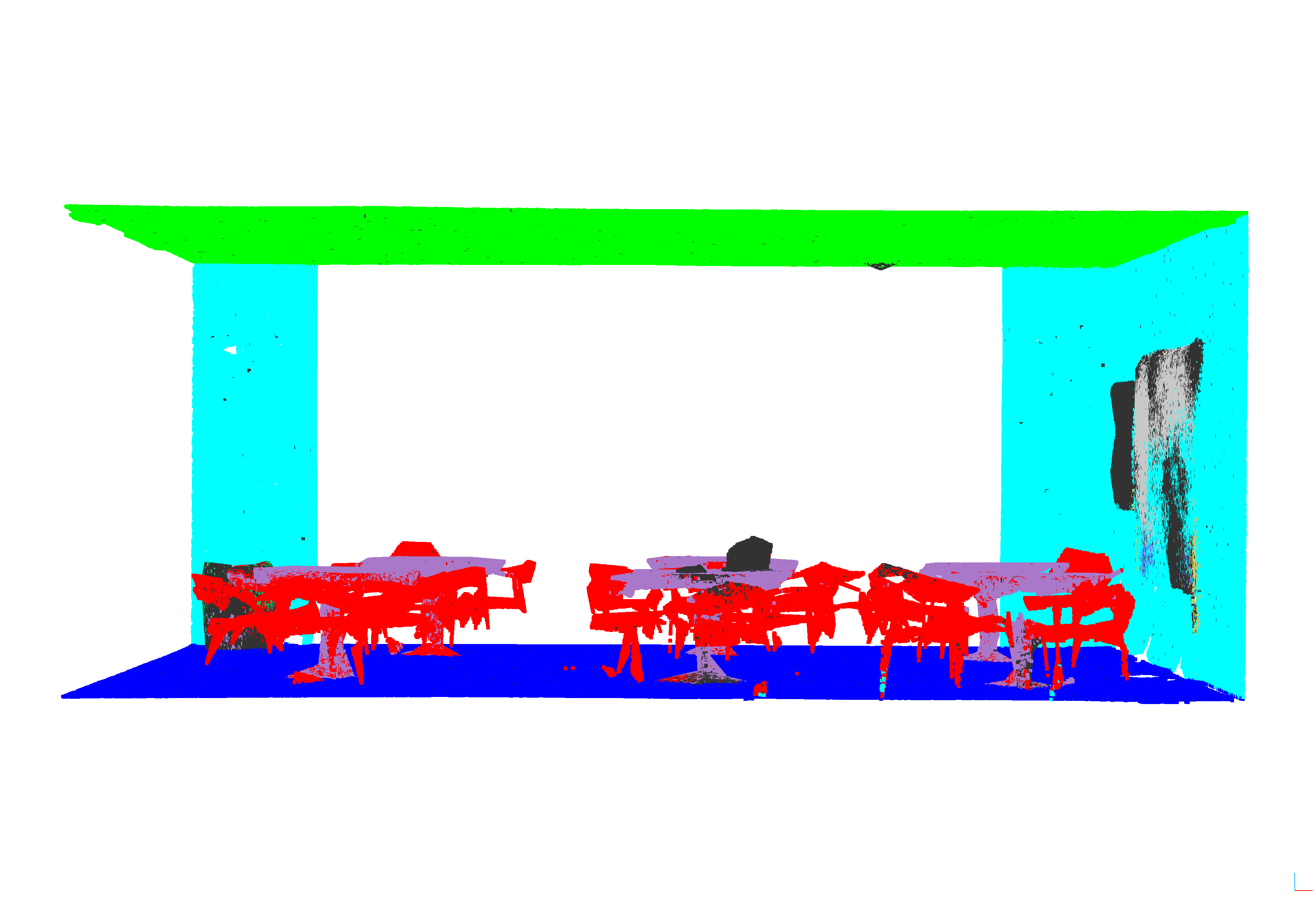} &
\includegraphics[width=0.45\linewidth]{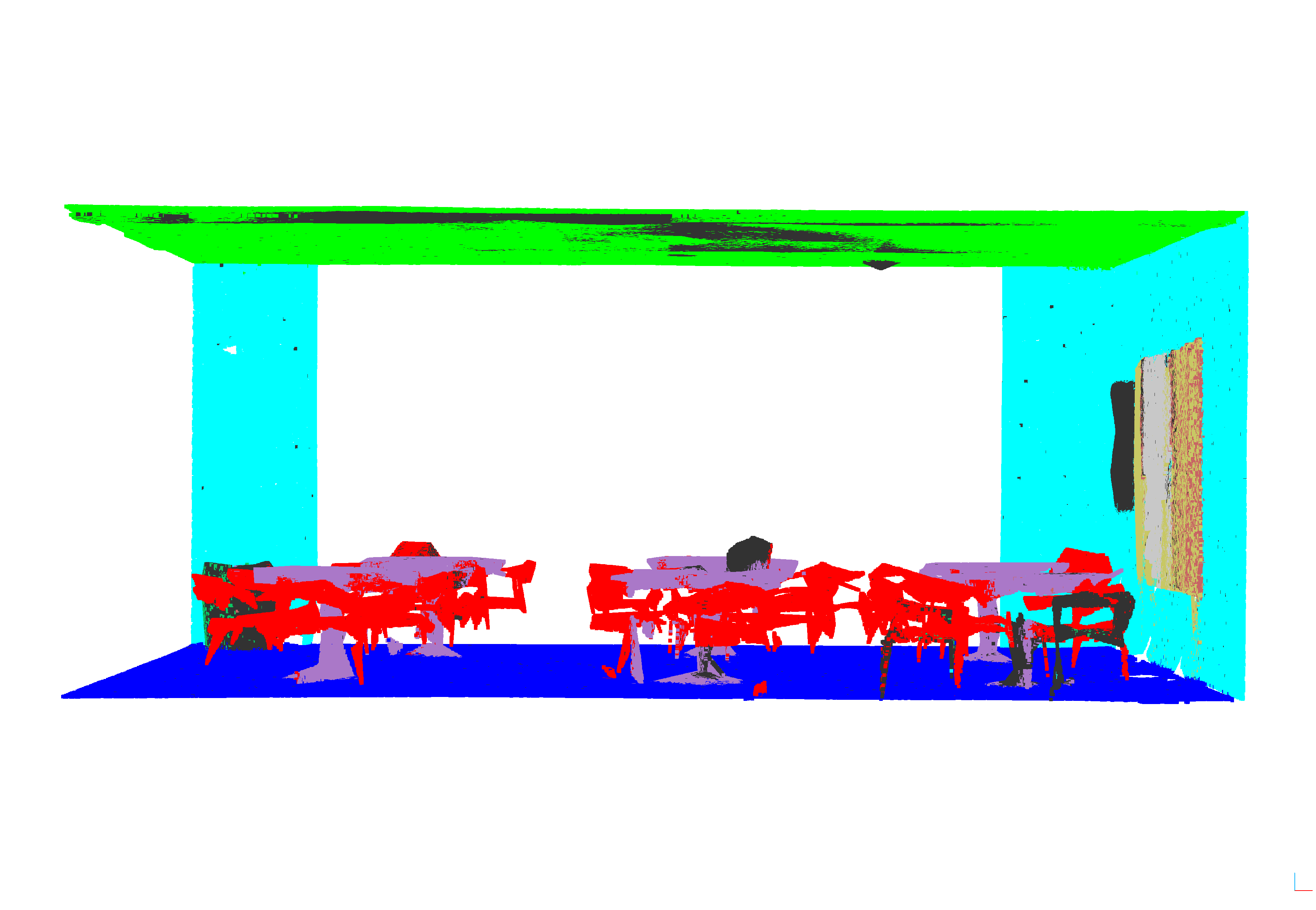} \\
\includegraphics[width=0.45\linewidth]{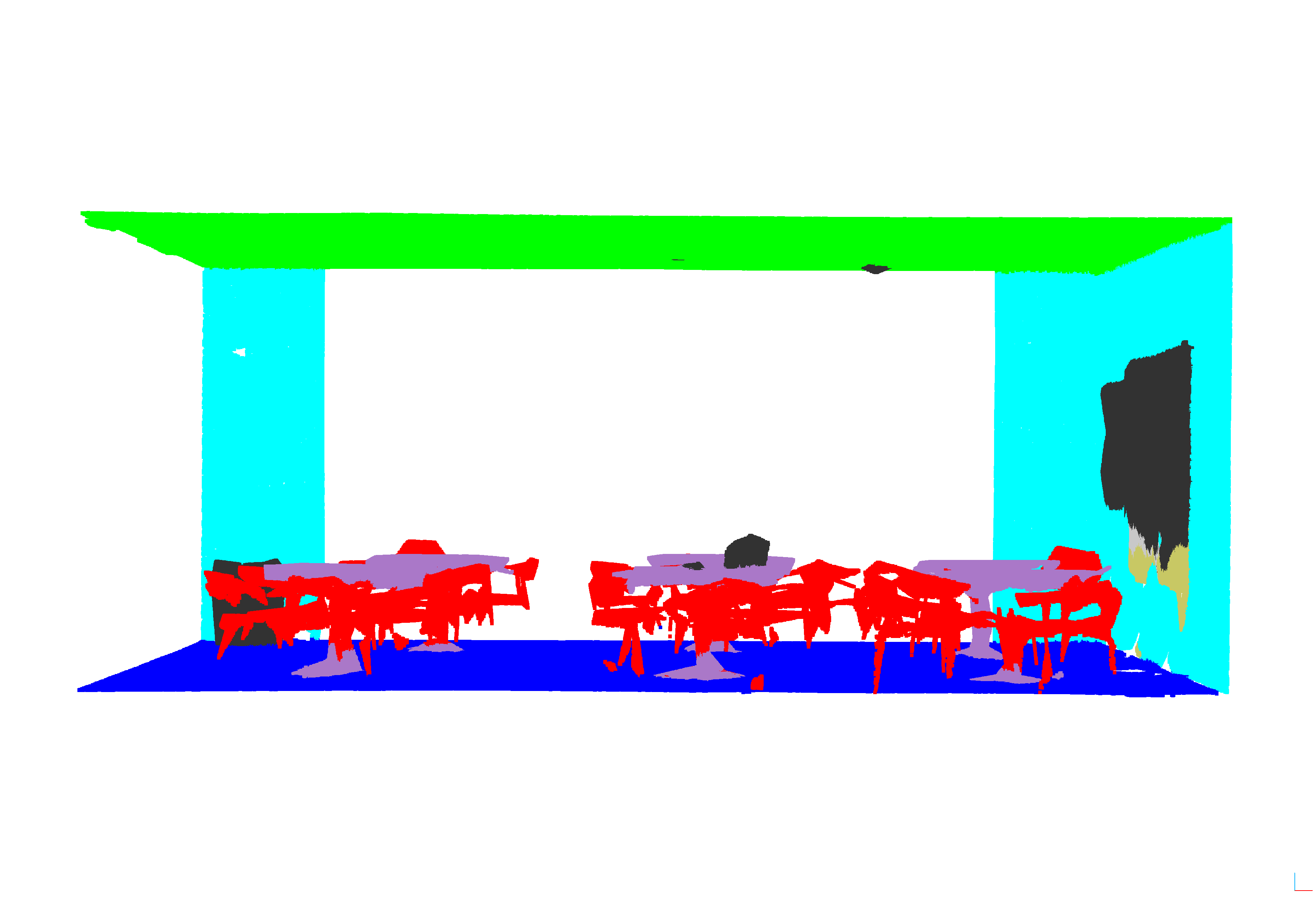} &
\includegraphics[width=0.45\linewidth]{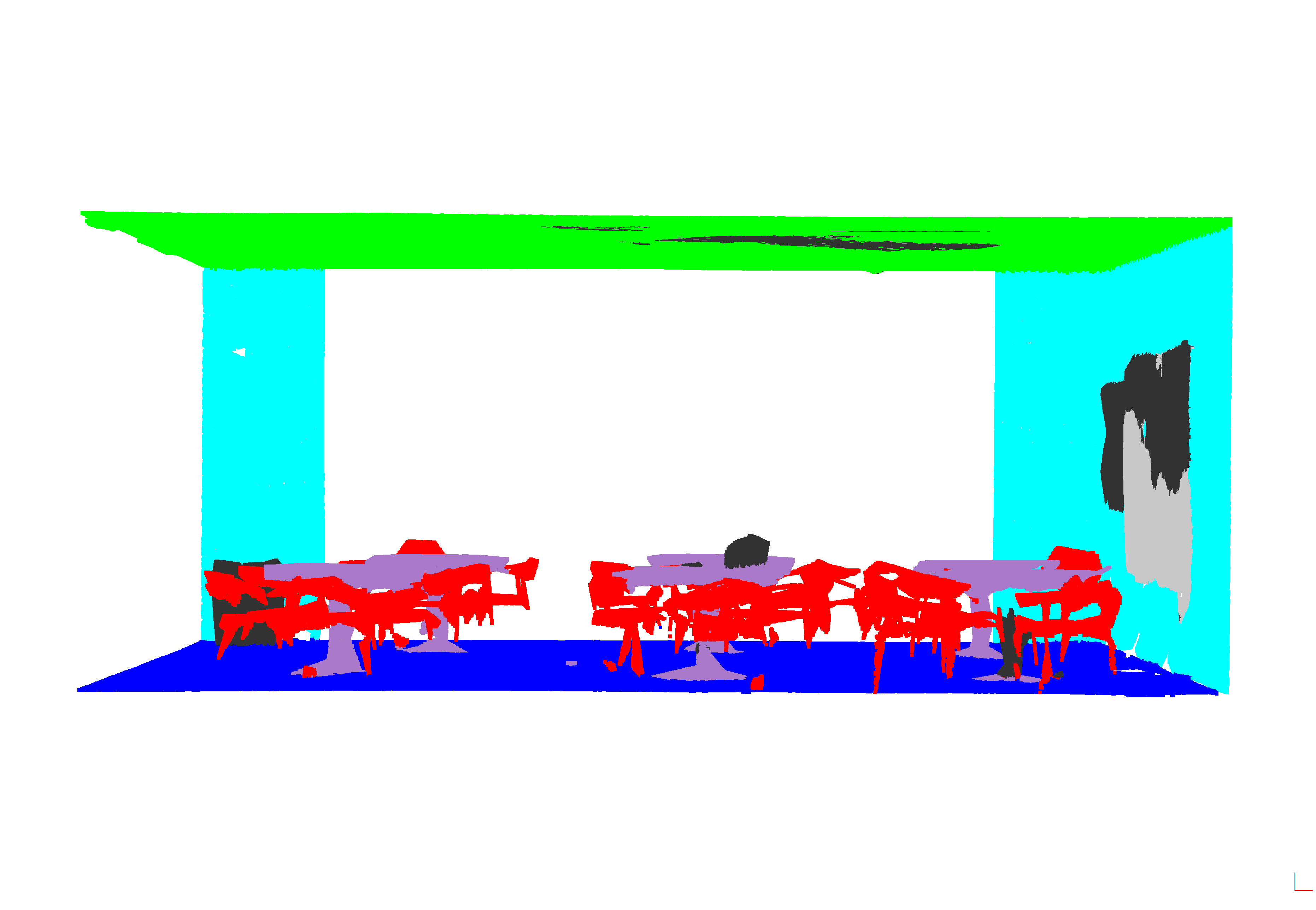} \\
\includegraphics[width=0.45\linewidth]{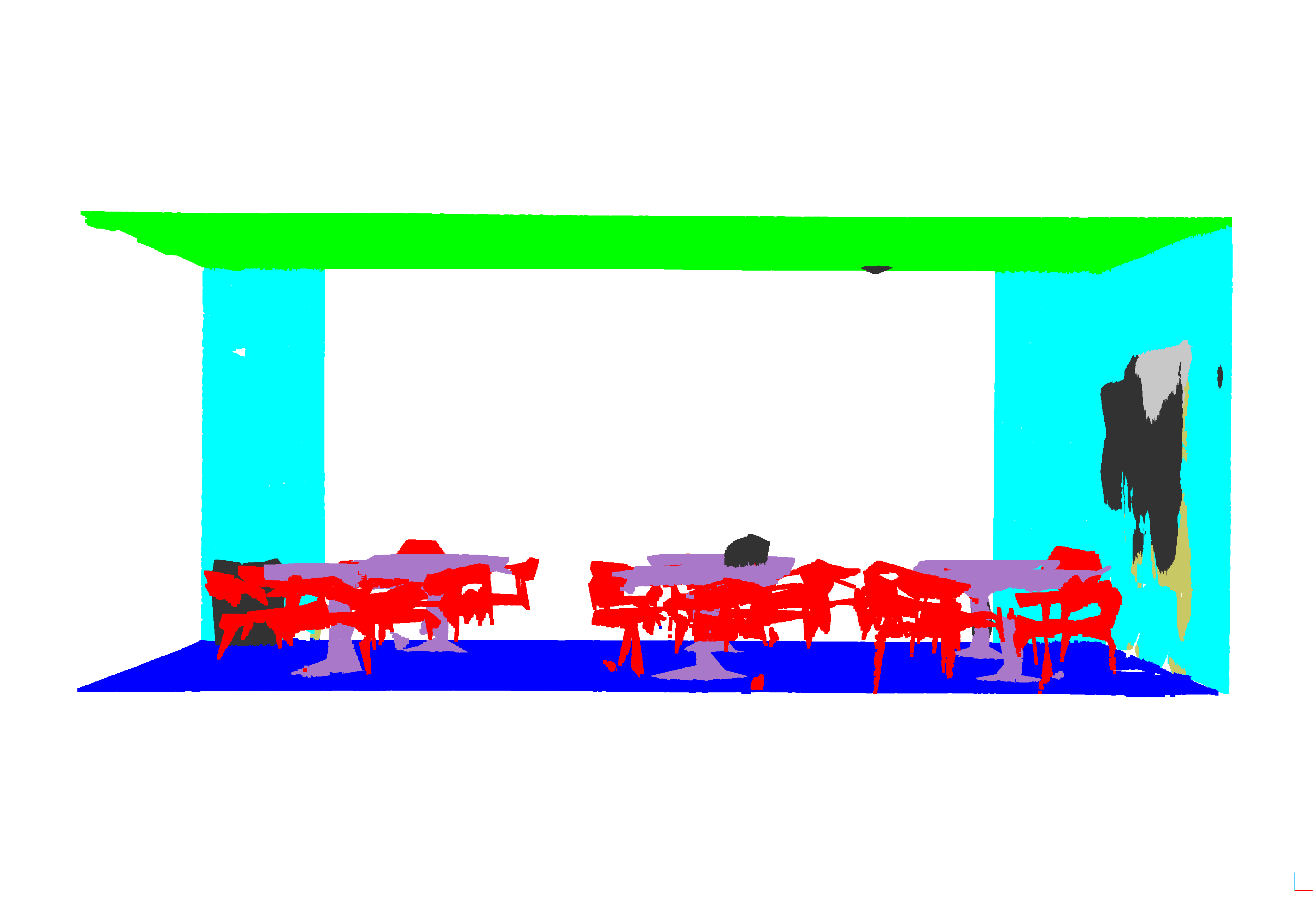} &
\includegraphics[width=0.45\linewidth]{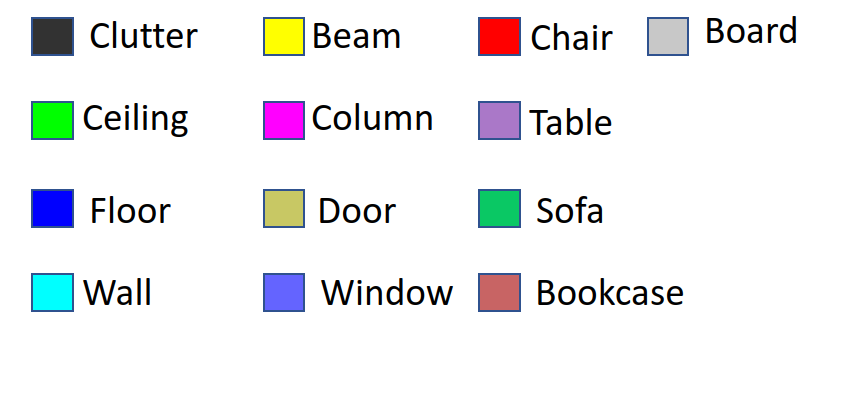}\\
\end{tabular}
\caption{Vizualizacija rezultata za sve modele i prostoriju \emph{lobby1} iz područja 5. Na slikama su redom (odozgo prema dolje, slijeva nadesno) prikazani: ulazni oblak točaka, stvarne oznake te oznake dobivene modelima \emph{PointCNN}, \emph{Cylinder3D}, \emph{PointNet++}, \emph{Point Transformer} i \emph{RepSurf}. }
\label{tab:confusion_matrices}
\end{figure*}

\begin{figure*}[h!]
\centering
\begin{tabular}{ll}
\includegraphics[width=0.45\linewidth]{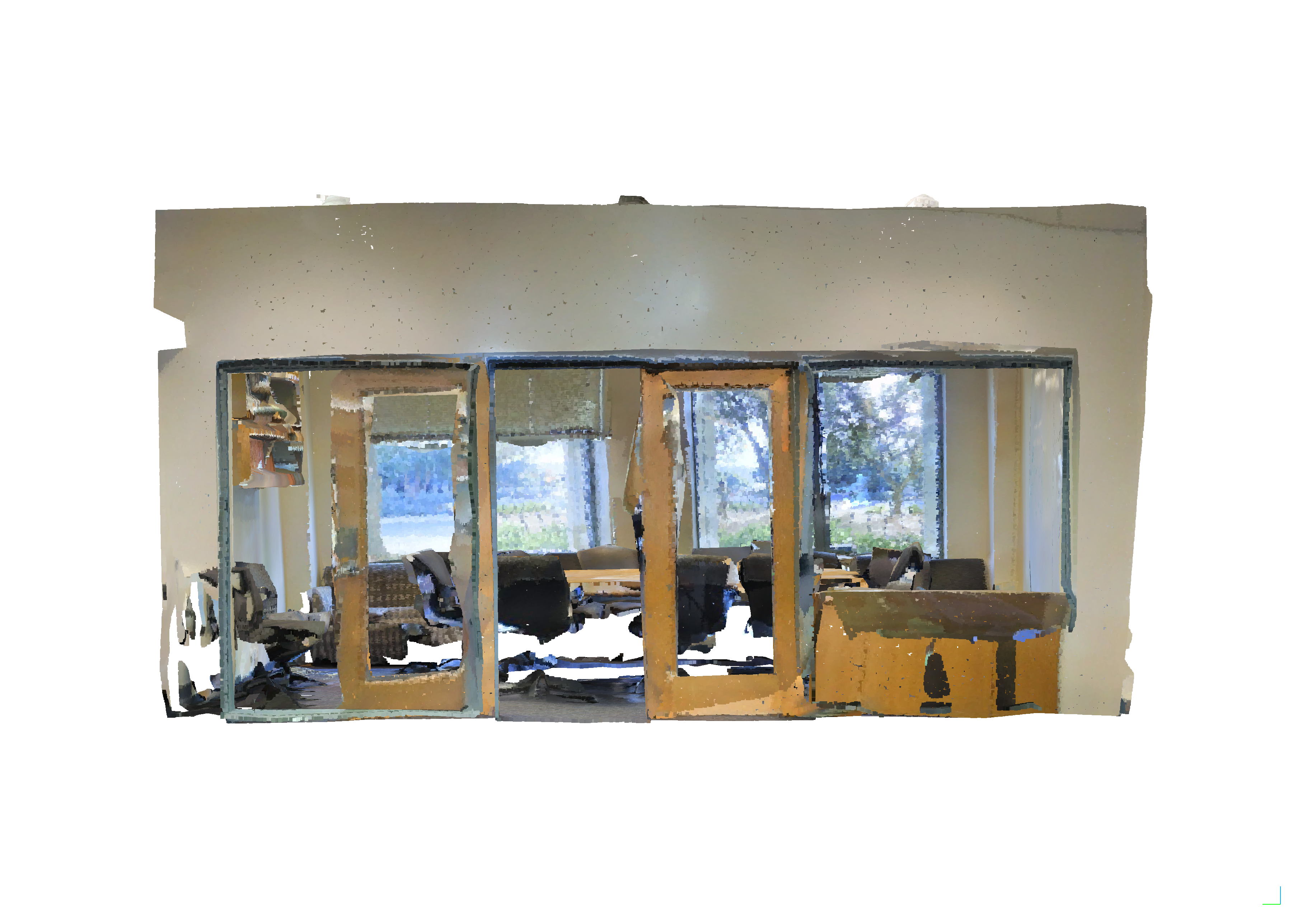} &
\includegraphics[width=0.45\linewidth]{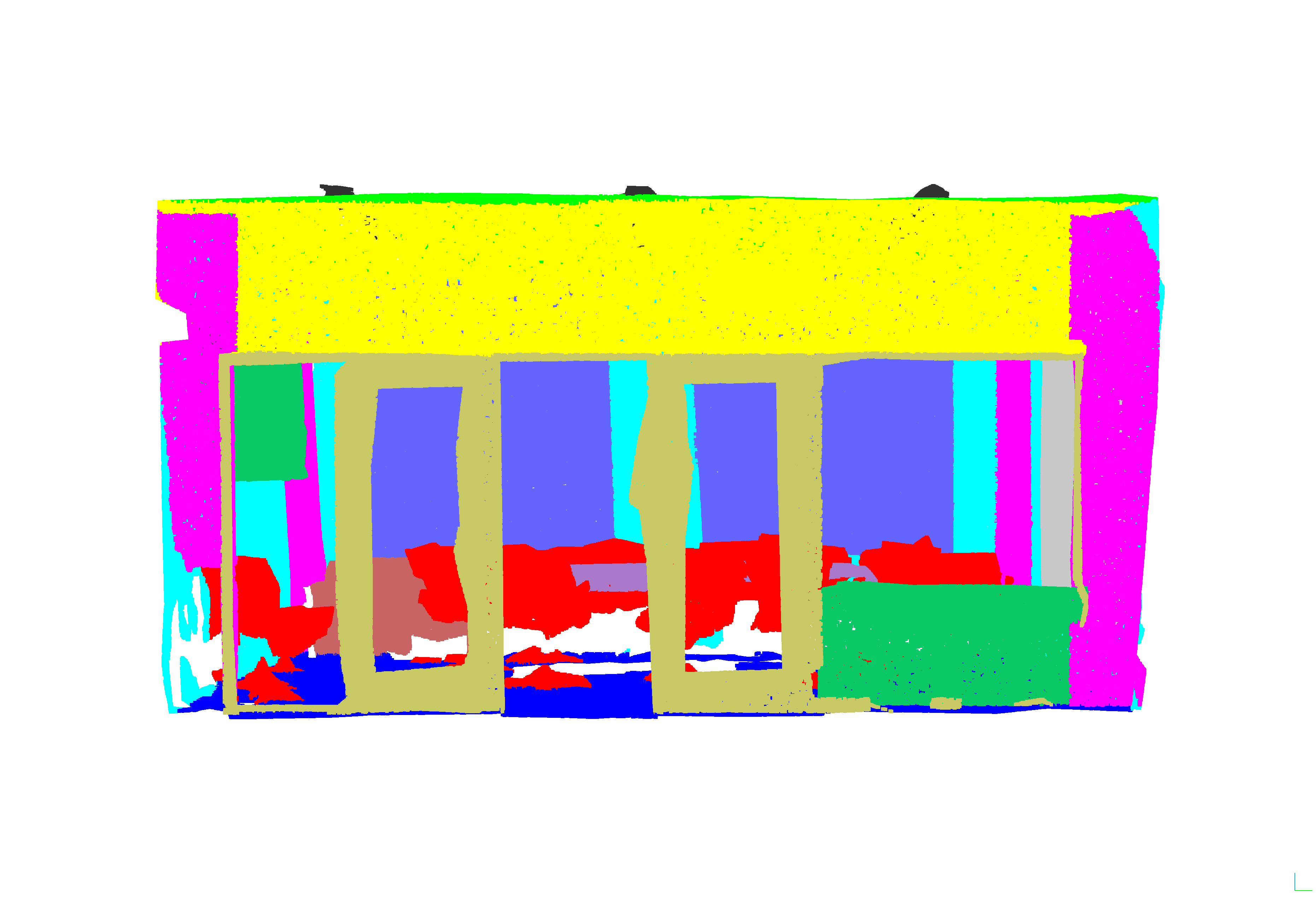} \\
\includegraphics[width=0.45\linewidth]{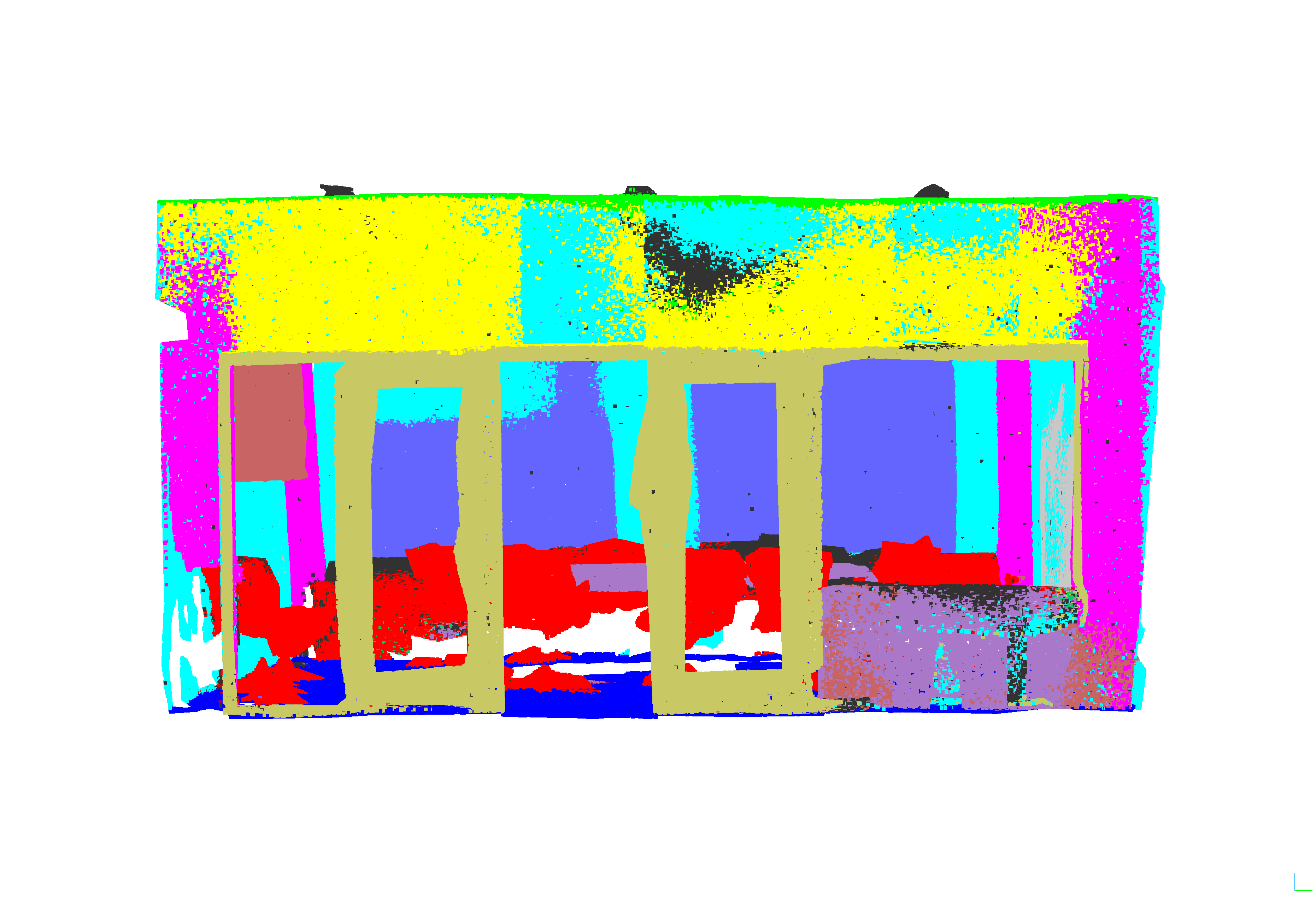} &
\includegraphics[width=0.45\linewidth]{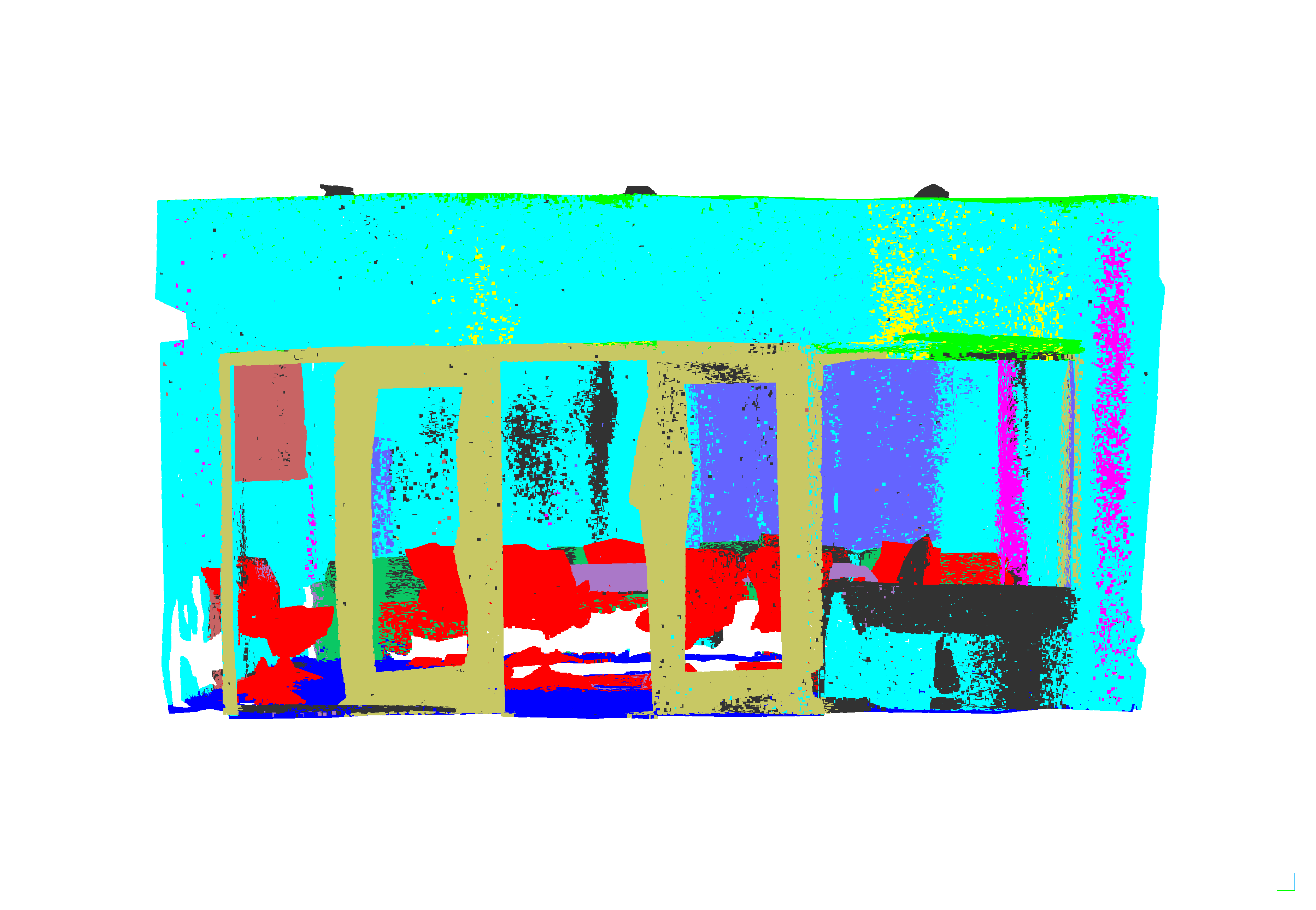} \\
\includegraphics[width=0.45\linewidth]{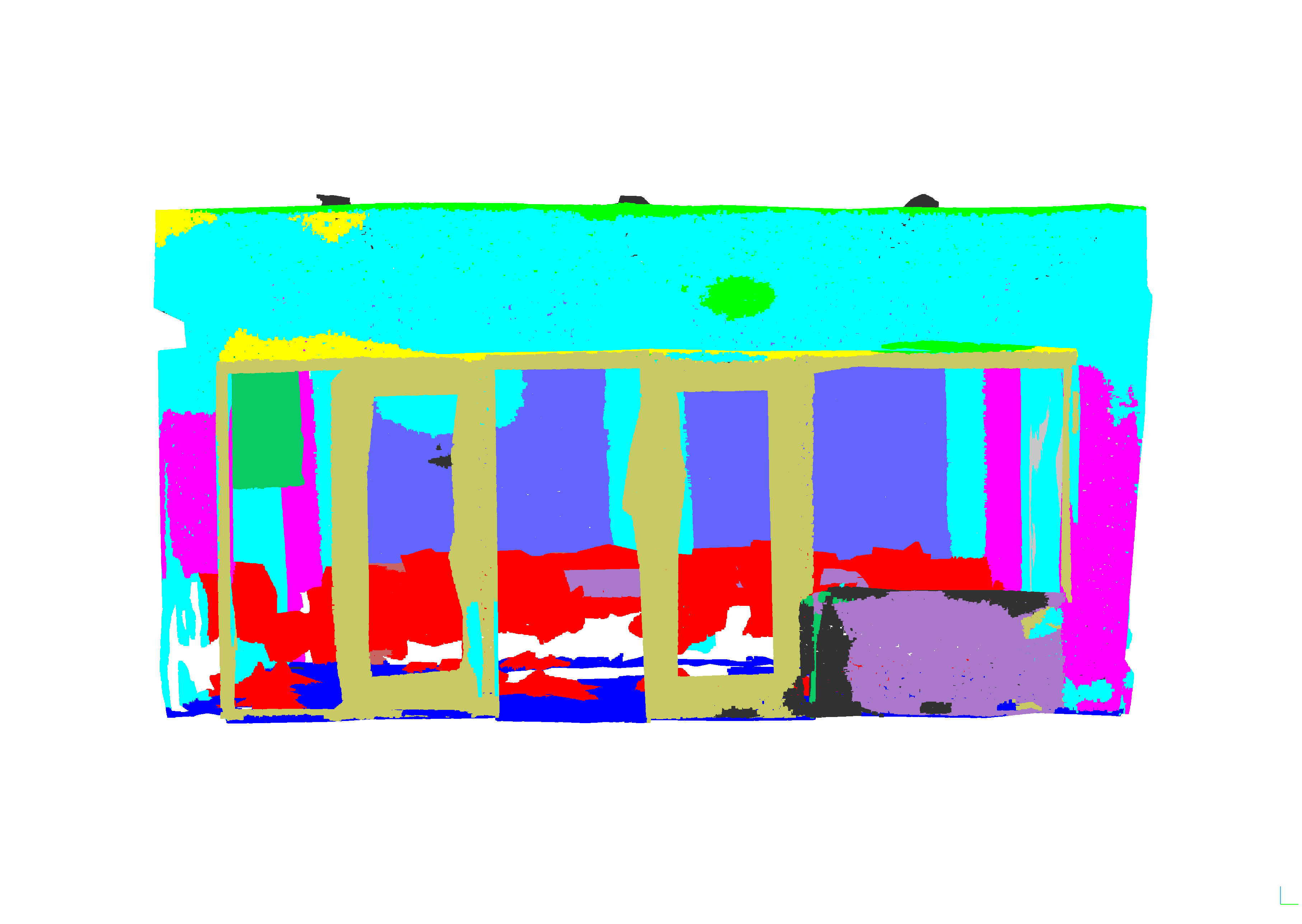} &
\includegraphics[width=0.45\linewidth]{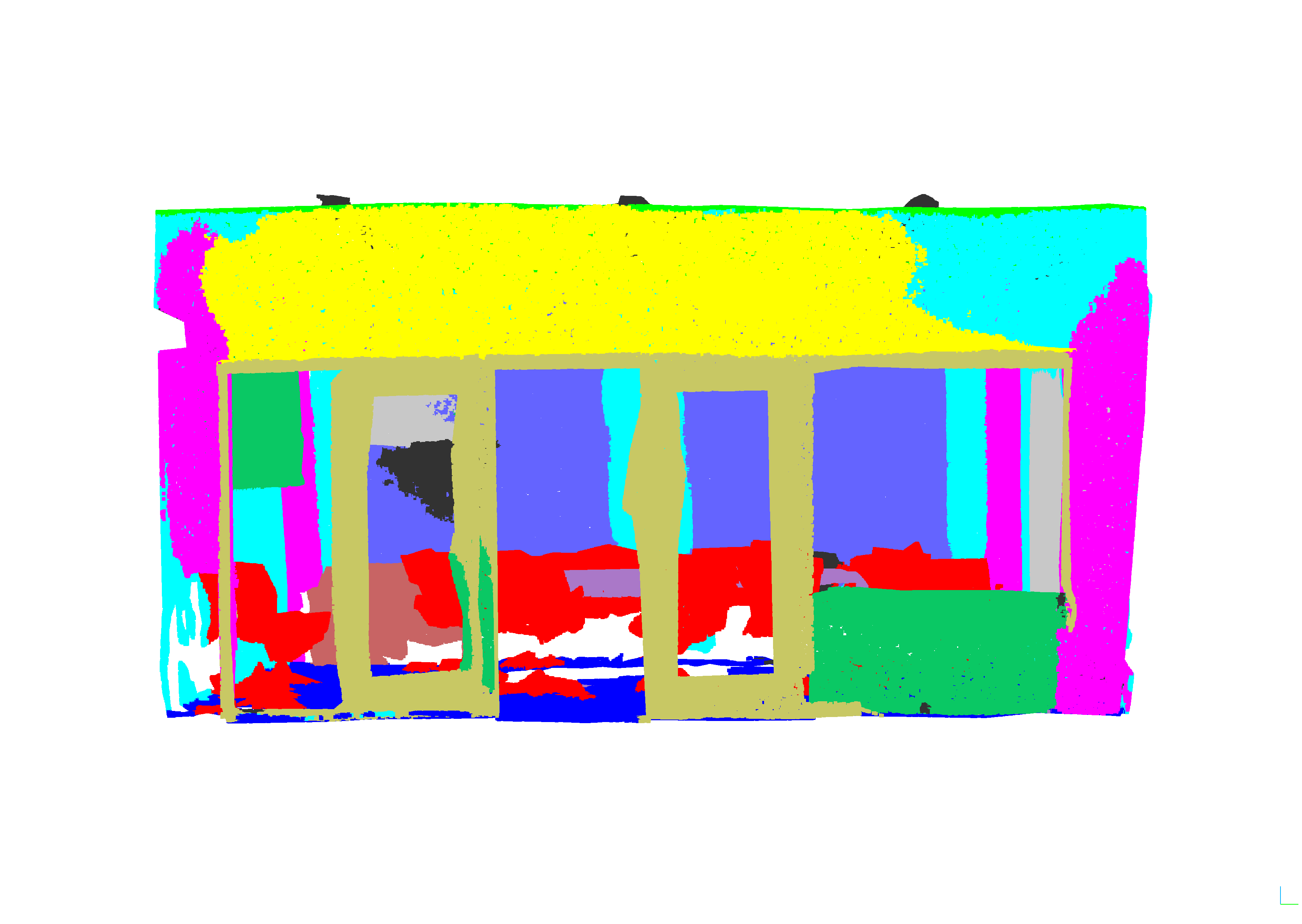} \\
\includegraphics[width=0.45\linewidth]{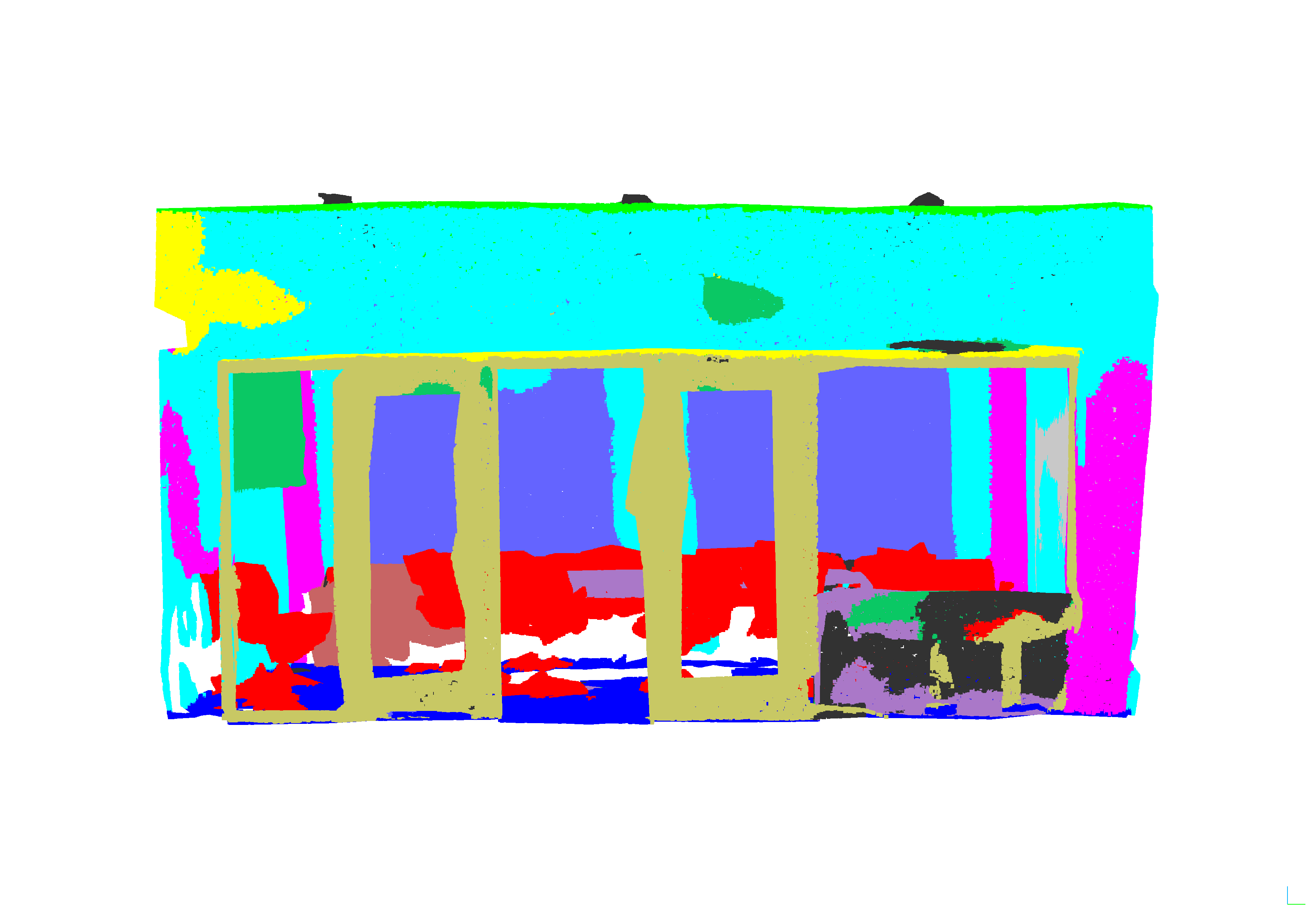} &
\includegraphics[width=0.45\linewidth]{img/colors_rep.png}\\
\end{tabular}
\caption{Vizualizacija rezultata za sve modele i prostoriju \emph{conferenceRoom2} iz područja 1. Na slikama su redom (odozgo prema dolje, slijeva nadesno) prikazani: ulazni oblak točaka, stvarne oznake te oznake dobivene modelima \emph{PointCNN}, \emph{Cylinder3D}, \emph{PointNet++}, \emph{Point Transformer} i \emph{RepSurf}. } 
\label{tab:confusion_matrices}
\end{figure*}

\end{document}